\pdfoutput=1

\documentclass[11pt]{article}

\usepackage{acl}

\usepackage{times}
\usepackage{latexsym}
\usepackage{graphicx}
\usepackage{amsmath}
\usepackage{amssymb}
\usepackage{multirow}
\usepackage{todonotes}

\usepackage{booktabs}
\usepackage[T1]{fontenc}

\usepackage[utf8]{inputenc}

\usepackage{enumitem}
\usepackage{microtype}

\usepackage{amsmath,amsfonts,bm}

\DeclareMathAlphabet{\mathsfit}{\encodingdefault}{\sfdefault}{m}{sl}
\SetMathAlphabet{\mathsfit}{bold}{\encodingdefault}{\sfdefault}{bx}{n}

\newcommand{\mypara}[1]{\noindent\textbf{#1}}
\newcommand\bfunder[1]{\textbf{\underline{#1}}}

\title{Paragraph-based Transformer Pre-training for Multi-Sentence Inference}

\author{Luca Di Liello$^{1}$\thanks{\ \ Work done as an intern at Amazon Alexa AI}\ , Siddhant Garg$^{2}$, Luca Soldaini$^{3}$\thanks{\ \ Work completed at Amazon Alexa AI}\ , Alessandro Moschitti$^{2}$\\
$^{1}$University of Trento , $^{2}$Amazon Alexa AI, $^{3}$Allen Institute for AI\\
\texttt{luca.diliello@unitn.it} \\ \texttt{\{sidgarg,amosch\}@amazon.com} \\
\texttt{lucas@allenai.org} \\
}

\begin{document}
\maketitle
\begin{abstract}
\vspace{-1em}
Inference tasks such as answer sentence selection (AS2) or fact verification are typically solved by fine-tuning transformer-based models as individual sentence-pair classifiers. Recent studies show that these tasks benefit from modeling dependencies across multiple candidate sentences jointly. In this paper, we first show that popular pre-trained transformers perform \emph{poorly} when used for fine-tuning on multi-candidate inference tasks. We then propose a new pre-training objective that models the paragraph-level semantics across multiple input sentences. Our evaluation on three AS2 and one fact verification datasets demonstrates the superiority of our pre-training technique over the traditional ones for transformers used as joint models for multi-candidate inference tasks, as well as when used as cross-encoders for sentence-pair formulations of these tasks.
\end{abstract}

\section{Introduction}
\vspace{-0.5em}
Pre-trained transformers ~\cite{devlin2019bert,liu2019roberta,clark2020electra} have become the de facto standard for several NLP applications, by means of fine-tuning on downstream data. 
The most popular architecture uses  self-attention mechanisms for modeling long range dependencies between compounds in the text, to produce deep contextualized representations of the input. 
There are several downstream NLP applications that require reasoning across multiple inputs candidates jointly towards prediction. Some popular examples include (i) Answer Sentence Selection (AS2)~\cite{garg2019tanda}, which is a Question Answering (QA) task that requires selecting the best answer from a set of candidates for a question; and (ii) Fact Verification~\cite{thorne-etal-2018-fever}, which reasons whether a claim is supported/refuted by multiple evidences. Inherently, these tasks can utilize information from multiple candidates (answers/evidences) to support the prediction of a particular candidate.

Pre-trained transformers such as BERT are used for these tasks as cross-encoders by setting them as sentence-pair classification problems, i.e, aggregating inferences independently over each candidate. Recent studies ~\cite{zhang-etal-2021-joint, tymoshenko-moschitti-2021-strong} have shown that these tasks benefit from encoding multiple candidates together, e.g., encoding five answer candidates per question in the transformer, so that the cross-attention can model dependencies between them. However, \citeauthor{zhang-etal-2021-joint} only improved over the pairwise cross-encoder by aggregating multiple pairwise cross-encoders together (one for each candidate), and not by jointly encoding all candidates together in a single model.

In this paper, we first show that popular pre-trained transformers such as RoBERTa perform \emph{poorly} when used for jointly modeling inference tasks (e.g., AS2) using multi-candidates. We show that this is due to a shortcoming of their pre-training objectives, being unable to capture meaningful dependencies among multiple candidates for the fine-tuning task. To improve this aspect, we propose a new pre-training objective for `joint' transformer models, which captures paragraph-level semantics across multiple input sentences. Specifically, given a target sentence $s$ and multiple sentences (from the same/different paragraph/document), the model needs to recognize which  sentences belong to the same paragraph as $s$ in the document used.

Joint inference over multiple-candidates entails modeling interrelated information between multiple \emph{short} sentences, possibly from different paragraphs or documents. This differs from related works~\cite{Beltagy2020Longformer,zaheer2020bigbird,DBLP:journals/corr/abs-2110-08499} that reduce the asymptotic complexity of transformer attention to model long contiguous inputs (documents) to get longer context for tasks such as machine reading and summarization.

We evaluate our pre-trained multiple-candidate based joint models by (i) performing AS2 on ASNQ~\cite{garg2019tanda}, WikiQA~\cite{yang2015wikiqa}, TREC-QA~\cite{wang-etal-2007-jeopardy} datasets; and (ii) Fact Verification on the FEVER~\cite{thorne-etal-2018-fever} dataset. We show that our pre-trained joint models substantially improve over the performance of transformers such as RoBERTa being used as joint models for multi-candidate inference tasks, as well as when being used as cross-encoders for sentence-pair formulations of these tasks. 

\section{Related Work}
\label{sec:related_work}

\mypara{Multi-Sentence Inference:} Inference over a set of multiple candidates has been studied in the past~\cite{10.1145/3132847.3133089,qingyao2018}. The most relevant for AS2 are the works of \citet{bonadiman-moschitti-2020-study} and \citet{zhang-etal-2021-joint}, the former improving over older neural networks but failing to beat the performance of transformers; the latter using task-specific models (answer support classifiers) on top of the transformer for performance improvements. For fact verification, ~\citet{tymoshenko-moschitti-2021-strong} propose jointly embedding multiple evidence with the claim towards improving the performance of baseline pairwise cross-encoder transformers.

\mypara{Transformer pre-training Objectives:} Masked Language Modeling (MLM) is a popular transformer pre-training objective~\cite{devlin2019bert,liu2019roberta}. Other models are trained using token-level~\cite{clark2020electra, joshi-etal-2020-spanbert, yang2020xlnet, diliello2021efficient} and/or sentence-level~\cite{devlin2019bert,lan2020albert,wang2019structbert} objectives. REALM~\cite{guu2020realm} uses a differentiable neural retriever over Wikipedia to improve MLM pre-training. This differs from our pre-training setting as it uses additional knowledge to improve the pre-trained LM. DeCLUTR~\cite{giorgi-etal-2021-declutr} uses a contrastive learning objective for cross-encoding two sentences coming from the same/different documents in a transformer. DeCLUTR is evaluated for sentence-pair classification tasks and embeds the two inputs independently without any cross-attention, which differs from our setting of embedding multiple candidates jointly for inference.

\mypara{Modeling Longer Sequences:} \citet{Beltagy2020Longformer,zaheer2020bigbird} reduce the asymptotic complexity of transformer attention to model very long inputs for longer context. For tasks with short sequence lengths, LongFormer works on par or slightly worse than RoBERTa (attributed to reduced attention computation). These works encode a single contiguous long piece of text, which differs from our setting of having multiple \emph{short} candidates, for a topic/query, possibly from different paragraphs and documents.
DCS~\cite{ginzburg-etal-2021-self} proposes a cross-encoder for the task of document-pair matching. DCS is related to our work as it uses a contrastive pre-training objective over two sentences extracted from the same paragraph, however different from our joint encoding of multiple sentences, DCS individually encodes the two sentences and then uses the InfoNCE loss over the embeddings. CDLM~\cite{caciularu-etal-2021-cdlm-cross} specializes the Longformer for document-pair matching and cross-document coreference resolution. While the pre-training objective in CDLM exploits information from multiple documents, it differs from our setting of joint inference over multiple short sentences.

\section{Multi-Sentence Transformers Models}
\label{sec:objectives}
\vspace{-0.25em}
\subsection{Multi-sentence Inference Tasks} 

\vspace{-0.25em}
\mypara{AS2:} We denote the question by $q$, and the set of answer candidates by $C{=}\{c_{1},\dots c_{n}\}$. The objective is to re-rank $C$ and find the best answer $A$ for q. AS2 is typically treated as a binary classification task: first, a model $f$ is trained to predict the correctness/incorrectness of each $c_i$; then, the candidate with the highest likelihood of being correct is selected as an answer, \textit{i.e.}, $A{=} \text{argmax}^n_{i=1} \text{ } f(c_i)$. Intuitively, modeling interrelated information between multiple $c_i$'s can help in selecting the best answer candidate \cite{zhang-etal-2021-joint}.

\mypara{Fact Verification:} We denote the claim by $F$, and the set of \emph{evidences} by $C{=}\{c_{1}\dots c_n\}$ that are retrieved using DocIR. The objective is to predict whether $F$ is supported/refuted/neither using $C$ (at least one evidence $c_i$ is required for supporting/refuting $F$). ~\citet{tymoshenko-moschitti-2021-strong} jointly model evidences for supporting/refuting a claim as they can complement each other.

\begin{figure}[t]
  \centering
  \includegraphics[width=\linewidth]{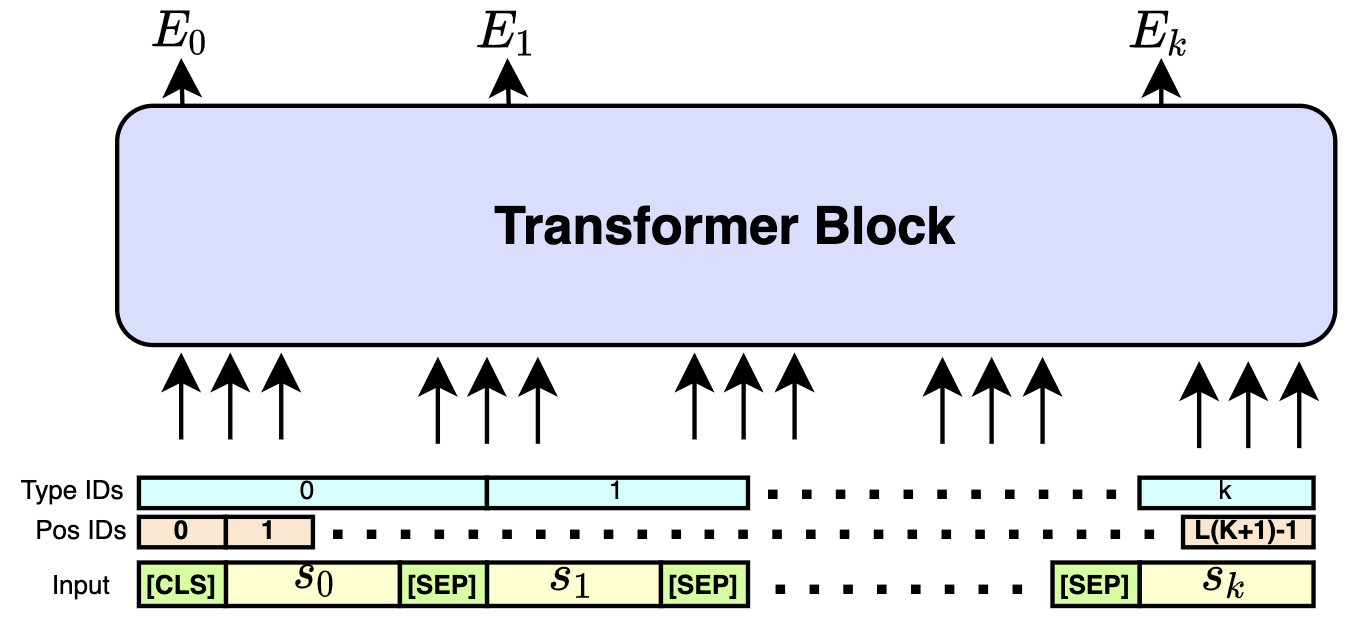}
  \caption{Multi-sentence `Joint' transformer model. $E_i$ refers to embedding for the question/each candidate.}
  \label{fig:joint_model}
  \vspace{.5em}
\end{figure}

\subsection{Joint Encoder Architecture} 
For jointly modeling multi-sentence inference tasks, we use a monolithic transformer cross-encoder to encode multiple sentences using self-attention as shown in Fig~\ref{fig:joint_model}. 
To perform joint inference over $k$ sentences for question $q$ or claim $F$, the model receives concatenated sentences $[s_0\dots s_k]$ as input, where the first sentence is either the question or the claim ($s_0{=}q$ or $s_0{=}F$), and the remainder are $k$ candidates $s_i{=}c_i \ ,i{=}\{1\dots k\}$. 
We pad (or truncate) each sentence $s_i$ to the same fixed length $L$ (total input length $L{\times}(k+1)$), and use the embedding for the {\small [CLS] / [SEP]} token in front of each sentence $s_i$ as its embedding (denoted by $E_i$). Similar to \citeauthor{devlin2019bert}, we create positional embeddings of tokens using integers $0$ to $L(k{+}1){-}1$, and extend the token type ids from $\{0,1\}$ to $\{0\dots k\}$ corresponding to $(k+1)$ input sentences.

\subsection{Inference using Joint Transformer Model} 
We use the output embeddings $[E_0\dots E_k]$ of sentences for performing prediction as following:

\mypara{Predicting a single label:} We use two separate classification heads to predict a single label for the input to the joint model $[s_0\dots s_k]$: (i) \textbf{IE$_1$:} a linear layer on the output embedding $E_0$ of $s_0$ (similar to BERT) referred to as the Individual Evidence (IE$_1$) inference head, and (ii) \textbf{AE$_1$:} a linear layer on the average of the output embeddings $[E_0,E_1,\dots,E_k]$ to explicitly factor in information from all candidates, referred to as the Aggregated Evidence (AE$_1$) inference head. For Fact Verification, we use prediction heads IE$_1$ and AE$_1$.

\begin{figure}[t]
  \centering
  \includegraphics[width=\linewidth]{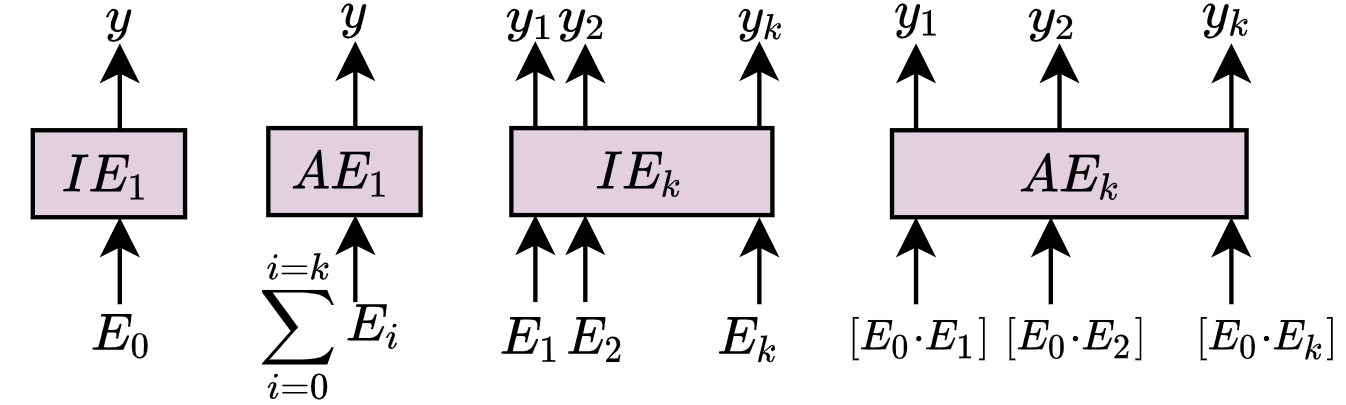}
  \vspace{-0.75em}
  \caption{Inference heads for joint transformer model. $E_i$ refers to embedding for the question/each candidate.}
  \label{fig:inference_heads}
\end{figure}

\mypara{Predicting Multiple Labels:} We use two separate classification heads to predict $k$ labels, one label each for every input $[s_1\dots s_k]$ specific to $s_0$: (i) \textbf{IE$_k$:} a shared linear layer applied to the output embedding $E_i$ of each candidate $s_i \ , \ i \in \{1\dots k\}$ referred to as $k$-candidate Individual Evidence (IE$_k$) inference head, and (ii) \textbf{AE$_k$:} a shared linear layer applied to the concatenation of output embedding $E_0$ of input $s_0$ and the output embedding $E_i$ of each candidate $s_i \ , \ i \in \{1\dots k\}$ referred to as $k$-candidate Aggregated Evidence (AE$_k$) inference head. For AS2, we use prediction heads IE$_k$ and AE$_k$. Prediction heads are illustrated in Figure~\ref{fig:inference_heads}.

\subsection{Pre-training with Paragraph-level Signals} 

Long documents are typically organized into paragraphs to address the document's general topic from different viewpoints. The majority of transformer pre-training strategies have not exploited this rich source of information, which can possibly provide some weak supervision to the otherwise unsupervised pre-training phase. To enable joint transformer models to effectively capture dependencies across multiple sentences, we design a new pre-training task where the model is (i) provided with $(k+1)$ sentences $\{s_0\dots s_k\}$, and (ii) tasked to predict which sentences from $\{s_1\dots s_k\}$ belong to the same paragraph $P$ as $s_0$ in the document $D$. We call this pre-training task Multi-Sentences in Paragraph Prediction (\textbf{MSPP}). We use the IE$_k$ and AE$_k$ prediction heads, defined above, on top of the joint model to make $k$ predictions $p_i$ 
corresponding to whether each sentence $s_i, \ i{\in}\{1\dots k\}$ lies in the same paragraph $P \in D$ as $s_0$. More formally: 

$  p_i = \begin{cases}
       1 \ \ \text{if} \ s_0, s_i \in P \ \text{in} \ D\\
       0 \ \ \text{otherwise} \\
     \end{cases} \forall i {=} \{1,\dots,k\}
$

\noindent We randomly sample a sentence from a paragraph $P$ in a document $D$ to be used as $s_0$, and then (i) randomly sample $k_1$ sentences (other than $s_0$) from $P$ as positives, (ii) randomly sample $k_2$ sentences from paragraphs other than $P$ in the same document $D$ as hard negatives, and (iii) randomly sample $k_3$ sentences from documents other than $D$ as easy negatives (note that $k_1 {+} k_2 {+} k_3 {=k}$). 

\section{Experiments}
\label{sec:experiments}
We evaluate our joint transformers on three AS2 and one Fact Verification datasets~\footnote{Code and pre-trained model checkpoints: \url{https://github.com/amazon-research/wqa-multi-sentence-inference}}.
Common LM benchmarks, such as GLUE~\cite{wang-etal-2018-glue}, are not suitable for our study as they only involve sentence pair classification.

\begin{table*}[t]
    \centering
    \resizebox{\linewidth}{!}{
    \begin{tabular}{lccccccccccc}
    \toprule
\multirow{2}{*}{\textbf{Model}} & \multicolumn{3}{c}{\textbf{ASNQ}} & &\multicolumn{3}{c}{\textbf{WikiQA}} & & \multicolumn{3}{c}{\textbf{TREC-QA}} \\ \cmidrule{2-4} \cmidrule{6-8} \cmidrule{10-12}
                       & \textbf{P@1}    & \textbf{MAP}    & \textbf{MRR}   & &\textbf{ P@1}     & \textbf{MAP}     & \textbf{MRR}   & & \textbf{P@1}   & \textbf{MAP}     & \textbf{MRR}     \\
    \midrule
\vspace{0.3em}
Pairwise RoBERTa-Base                                           & 61.8 (0.2) & 66.9 (0.1) & 73.1 (0.1)  & & 77.1 (2.1)  & 85.3 (0.9)    & 86.5 (1.0)            & & 87.9 (2.2)   & 89.3 (0.9)   & 93.1 (1.0)   \\
\vspace{0.3em}
Joint RoBERTa-Base ${\rightarrow}$ FT IE$_k$                    & 3.4 (2.3)  & 8.0 (1.9)  & 10.0 (2.4)  & & 19.7 (1.9)  & 39.4 (1.6)    & 40.3 (1.8)            & & 30.9 (5.4)   & 41.9 (2.4)   & 50.8 (3.9)   \\
\vspace{0.3em}
Joint RoBERTa-Base ${\rightarrow}$ FT AE$_k$                    & 3.6 (2.7)  & 8.0 (2.2)  & 10.2 (2.8)  & & 18.7 (3.9)  & 39.0 (2.8)    & 39.7 (2.9)            & & 29.7 (6.9)   & 42.3 (3.2)   & 49.2 (5.0)   \\
\vspace{0.3em}
(\textbf{Ours}) Joint MSPP IE$_k$ ${\rightarrow}$ FT IE$_k$     & \bfunder{63.0} (0.3) & 67.2 (0.2) & \textbf{73.7} (0.2) & & \bfunder{82.7} (2.2) & \bfunder{88.5} (1.5)  & \bfunder{89.0} (1.5)  && \bfunder{91.7} (2.2) & \bfunder{91.1} (0.5) & \bfunder{95.2} (1.3) \\
\vspace{0.3em}
(\textbf{Ours}) Joint MSPP AE$_k$ ${\rightarrow}$ FT AE$_k$     & \bfunder{63.0} (0.3) & \textbf{67.3} (0.2) & \textbf{73.7} (0.2) & & \underline{81.9} (2.6) & \underline{87.9} (1.4) & \underline{89.0} (1.5)  & & 88.7 (0.8) & 90.1 (1.0)  & 93.6 (0.6)   \\
\bottomrule
\end{tabular} }
\vspace{0.2em}
    \caption{Results (std. dev. in parenthesis) on AS2 datasets. MSPP, FT refer to our pre-training task and fine-tuning respectively. We indicate the prediction head (IE$_k$/AE$_k$) used for both pre-training and fine-tuning. We underline statistically significant gains over the baseline (Student \textit{t}-test with $95\%$ confidence level).}
    \label{tab:results_as2}
\vspace{0.4em}
\end{table*}


\subsection{Datasets}
\label{subsec:datasets}

\mypara{Pre-training: } To eliminate any improvements stemming from usage of more data, we perform pre-training on the same corpora as RoBERTa: English Wikipedia, the BookCorpus, OpenWebText and CC-News. 
For our proposed pre-training, we randomly sample sentences from paragraphs as $s_0$, and choose $k_1{=}1, k_2{=}2, k_3{=}2$ as the specific values for creating positive and negative candidates for $s_0$. For complete details refer to Appendix~\ref{app:datasets}.

\mypara{Fine-tuning:} 
For AS2, we compare performance with MAP, MRR and Precision of top ranked answer (P@1). For fact verification, we measure Label Accuracy (LA). Brief description of datasets is presented below (details in Appendix~\ref{app:datasets}):
\vspace{-.5em}
\begin{itemize}[wide, labelwidth=!, labelindent=0pt]
\itemsep-0.25em 
    \item \textbf{ASNQ:} A large AS2 dataset~\cite{garg2019tanda} derived from NQ~\citep{kwiatkowski-etal-2019-natural},  where the candidate answers are from Wikipedia pages and the questions are from search queries of the Google search engine. We use the dev. and test splits released by ~\citeauthor{soldaini-moschitti-2020-cascade}. 
    \item \textbf{WikiQA:} An AS2 dataset~\cite{yang2015wikiqa} where the questions are derived from query logs of the Bing search engine, and the answer candidate are extracted from Wikipedia. We use the most popular \emph{clean} setting (questions having at least one positive and one negative answer). 
    \item \textbf{TREC-QA:} A popular AS2 dataset~\cite{wang-etal-2007-jeopardy} containing factoid questions. We only retain questions with at least one positive and one negative answer in the development and test sets.
    \item \textbf{FEVER:} A dataset for fact extraction and verification~\cite{thorne-etal-2018-fever} to retrieve evidences given a claim and identify if the evidences support/refute the claim. As we are interested in the fact verification sub-task, we use evidences retrieved by \citeauthor{liu2020kernel} using a BERT-based DocIR.
\end{itemize}

\subsection{Experimental Details and Baselines}

We use $k{=}5$ for our experiments (following \cite{zhang-etal-2021-joint} and \cite{tymoshenko-moschitti-2021-strong}), and perform continued pre-training starting from RoBERTa-Base using a combination of MLM and our MSPP pre-training for 100k steps with a batch size of 4,096. We use two different prediction heads, IE$_k$ and AE$_k$, for pre-training. For evaluation, we fine-tune all models on the downstream AS2 and FEVER datasets using the corresponding IE$_k$ and AE$_k$ prediction heads. We consider the pairwise RoBERTa-Base cross-encoder and RoBERTa-Base LM used as a joint model with IE$_k$ and AE$_k$ prediction heads as the baseline for AS2 tasks. For FEVER, we use several baselines: GEAR~\cite{zhou-etal-2019-gear}, KGAT~\cite{liu2020kernel}, Transformer-XH~\cite{zhao2020transformer-xh}, and three models from \cite{tymoshenko-moschitti-2021-strong}: (i) Joint RoBERTa-Base with IE$_1$ prediction head, (ii) Pairwise RoBERTa-Base with max-pooling, and (iii) weighted-sum heads. For complete experimental details, refer to Appendix~\ref{app:experiments}.

\subsection{Results}
\label{subsec:results}

\begin{table}[t]
    \centering
    \resizebox{\linewidth}{!}{
    \begin{tabular}{lccc}
    \toprule
    \textbf{Model} & \textbf{ASNQ} & \textbf{WikiQA} & \textbf{TREC-QA} \\
    \midrule
Pairwise RoBERTa-Base                                       & 61.8 (0.2)                & 77.1 (2.1)                & 87.9 (2.2) \\
Joint RoBERTa-Base ${\rightarrow}$ FT IE$_k$                & 25.2 (3.1)                & 24.6 (3.1)                & 57.6 (4.8) \\
Joint RoBERTa-Base ${\rightarrow}$ FT AE$_k$                & 25.4 (3.3)                & 26.4 (2.2)                & 60.9 (4.9) \\
(\textbf{Ours}) Joint MSPP IE$_k$ ${\rightarrow}$ FT IE$_k$ & \underline{63.9} (0.8)    & \bfunder{82.7} (3.0)      & \bfunder{92.2} (0.8) \\
(\textbf{Ours}) Joint MSPP AE$_k$ ${\rightarrow}$ FT AE$_k$ & \bfunder{64.3} (1.1)      & \underline{82.1} (1.1)    & 91.2 (2.9) \\
\bottomrule
\end{tabular} }
\vspace{0.2em}
    \caption{P@1 of joint models for AS2 when re-ranking answers ranked in top-5 by pairwise RoBERTa-Base. Statistically significant results (Student \textit{t}-test $95\%$) are underlined. Complete results in Appendix~\ref{app:complete_results}.}
    \label{tab:results_as2_top5}
\vspace{0.2em}
\end{table}

\mypara{Answer Sentence Selection:} The results for AS2 tasks are presented in Table~\ref{tab:results_as2}, averaged across five independent runs. From the table, we can see that the RoBERTa-Base when used as a joint model for multi-candidate inference using either the IE$_k$ or AE$_k$ prediction heads performs inferior to RoBERTa-Base used as a pairwise cross-encoder. Across five experimental runs, we observe that fine-tuning RoBERTa-Base as a joint model faces convergence issues (across various hyper-parameters) indicating that the MLM pre-training task is not sufficient to learn text semantics which can be exploited for multi-sentence inference. 

Our MSPP pre-trained joint models (with both IE$_k$, AE$_k$ heads) get significant improvements over the pairwise cross-encoder baseline and very large improvements over the RoBERTa-Base joint model. The former highlights modeling improvements stemming from joint inference over multiple-candidates, while the latter highlights improvements stemming from our MSPP pre-training strategy. Across all three AS2 datasets, our joint models are able to get the highest P@1 scores while also improving the MAP and MRR metrics. 

To demonstrate that our joint models can effectively use information from multiple candidates towards prediction, we perform a study in Table~\ref{tab:results_as2_top5} where the joint models are used to re-rank the top-$k$ candidates ranked by the pairwise RoBERTa-Base cross-encoder. Our joint models can significantly improve the P@1 over the baseline for all datasets. The performance gap stems from questions for which the pairwise RoBERTa model was unable to rank the correct answer at the top position, but support from other candidates in the top-k helped the joint model rank it in the top position. 

\begin{table}[t]
    \centering
    \resizebox{\linewidth}{!}{
    \begin{tabular}{lcccc}
    \toprule
        \textbf{Model} & \textbf{Dev} & \textbf{Test} \\
        \midrule
        GEAR & 70.69 & 71.60 \\
        KGAT with RoBERTa-Base & 78.29 & 74.07 \\
        Transformer-XH & 78.05 & 72.39 \\
        Pairwise RoBERTa-Base + MaxPool & 79.82 & - \\
        Pairwise RoBERTa-Base + WgtSum & 80.01 & - \\
        Joint RoBERTa-Base + FT IE$_1$ & 79.25 & 73.56 \\
        (\textbf{Ours}) Joint Pre IE$_k$ + FT IE$_1$  & \bfunder{81.21} (0.24)    & \textbf{74.39} \\
        (\textbf{Ours}) Joint Pre IE$_k$ + FT AE$_1$  & \underline{81.10} (0.15)  & 74.25 \\
        (\textbf{Ours}) Joint Pre AE$_k$ + FT IE$_1$  & \underline{81.18} (0.14)  & 73.77  \\
        (\textbf{Ours}) Joint Pre AE$_k$ + FT AE$_1$  & \bfunder{81.21} (0.16)    & 74.13 \\
        \bottomrule
    \end{tabular}
    }
\vspace{0.2em}
    \caption{Results on FEVER dev and test sets. For our method, prediction heads (IE$_1$/AE$_1$) are only used for fine-tuning (FT), while for pre-training (Pre) we use (IE$_k$/AE$_k$) heads. `-' denotes models not released publicly, and results not reported in the paper. Statistically significant results (Student \textit{t}-test $95\%$) are underlined.}
    \label{tab:results_fever}
    \vspace{.5em}
\end{table}

\mypara{Fact Verification:} The results for the FEVER task are presented in Table~\ref{tab:results_fever} and show that our joint models (pre-trained with both the IE$_k$ and AE$_k$ heads and fine-tuned with the IE$_1$ and AE$_1$ heads) outperform all previous baselines considered, including the RoBERTa-Base joint model directly applied for multi-sentence inference. 

\begin{table}[t!]
    \centering
    \resizebox{\linewidth}{!}{
    \begin{tabular}{l}
    \toprule
    \textbf{ASNQ} \\
    \midrule
    \textbf{Q:} Who invented the submarine during the civil war?\\
    \textbf{A1:} {\color{red} H.L. Hunley , often referred to as Hunley , was a submarine of the Confedera}\\
    \textbf{A2:} Hunley , McClintock , and Baxter Watson first built Pioneer , which was tested \\ in February  1862 in the Mississippi River and was later towed to Lake Pontchartrain \\ for additional trials .\\
    \textbf{A3:} {\color{ForestGreen} She was named for her inventor, Horace Lawson Hunley , shortly after she was} \\ {\color{ForestGreen} taken into government service under the control of the Confederate States Army} \\
    {\color{ForestGreen} at Charleston , South Carolina.} \\
    \textbf{A4:} 1864 painting of H.L. Hunley by Conrad Wise Chapman History Confederate States \\ Name : H.L. Hunley Namesake : Horace Lawson Hunley Builder : James McClintock \\ Laid down : Early 1863  Launched : July 1863 Acquired : August 1863 In service: Feb- \\ ruary 17 , 1864 Out of service : February 17, 1864 Status : Awaiting conservation General \\ characteristics Displacement : 7.5 short tons ( 6.8 metric tons ) Length : 39.5 ft \\
    \textbf{A5:} Johan F. Carlsen was born in Ærøskøbing April 9, 1841. \\
    \midrule
    \textbf{WikiQA} \\
    \midrule
    \textbf{Q:} What is the erb/heart?\\
    \textbf{A1:} {\color{red} Heart valves are labeled with "B", "T", "A", and "P".First heart sound: caused by} \\ { \color{red} atrioventricular valves - Bicuspid/Mitral (B) and Tricuspid (T).}\\
    \textbf{A2:} {\color{ForestGreen} Second heart sound caused by semilunar valves -- Aortic (A) and Pulmonary/} \\ {\color{ForestGreen} Pulmonic (P).}\\
    \textbf{A3:} Front of thorax , showing surface relations of bones , lungs (purple), pleura (blue), \\ and heart (red outline). \\
    \textbf{A4:} In cardiology, Erb's point refers to the third intercostal space on the left sternal \\ border where sS2 is best auscultated . \\
    \textbf{A5:} It is essentially the same location as what is referred to with left lower sternal \\ border (LLSB).\\
    \midrule
    \textbf{TREC-QA} \\
    \midrule
    \textbf{Q:} When was the Khmer Rouge removed from power ?\\
    \textbf{A1:} {\color{red} Sihanouk was named head of state after the Khmer Rouge seized power in 1975,} \\ {\color{red} but was locked in his palace by the communists as they embarked on their brutal}\\ {\color{red} attempt to create an agrarian utopia .}\\
    \textbf{A2:} When a Vietnamese invasion drove the Khmer Rouge from power in 1979, \\ Duch fled with other Khmer Rouge leaders into the jungles.\\
    \textbf{A3:} {\color{ForestGreen} Religious practices were revived after the Khmer Rouge were driven from power} \\ {\color{ForestGreen}  by a Vietnamese invasion in 1979}  \\
    \textbf{A4:} Moreover, 20 years after the Khmers Rouges were ousted from power, Cambodia \\ still struggles on the brink of chaos , ruled by the gun , not by law . \\
    \textbf{A5:} Sihanouk resigned in 1976 , but the Khmer Rouge kept him under house arrest \\ until they were driven from power by an invading Vietnamese army in 1979 .\\
    \bottomrule
    \end{tabular}}
\vspace{0.2em}
    \caption{Examples from AS2 datasets where the pairwise RoBERTa-Base model is unable to rank a {\color{ForestGreen} correct answer} for the question at the top position, but our joint model (Joint MSPP IE$_k$ ${\rightarrow}$ FT IE$_k$) can. We present answers $\{A1,\dots,A5\}$ in their ranked order by the pairwise RoBERTa-Base model. For all these examples we highlight the top ranked answer by the pairwise RoBERTa-Base model in {\color{red} red} since it is incorrect.}
    \label{tab:qualitative}
\end{table}

\mypara{Compute Overhead:} We present a \emph{simplified} latency analysis for AS2 (assuming sentence length $L$) as follows: a pairwise cross-encoder uses $k$ transformer steps with input length $2L$, while our model uses $1$ step with input length $(k{+}1){\times}L$. Since transformer attention scales quadratic on input length, our model should take $\frac{(k{+}1)^2}{4k}$ times the inference time of the cross-encoder, which is $1.8$ when $k{=}5$. However, when we fine-tune for WikiQA on one A100-GPU, we only observe latency increasing from $71s {\rightarrow} 81s$ (only $14.1\%$ increase). The input embeddings and feedforward layers vary linearly with input length, reducing overheads of self-attention. Refer to Appendix~\ref{app:number_params} for details.

\mypara{Qualitative Examples:}
We present some qualitative examples from the three AS2 datasets highlighting cases where the pairwise RoBERTa-Base model is unable to rank the correct answer on the top position, but our pre-trained joint model (Joint MSPP IE$_k$ ${\rightarrow}$ FT IE$_k$) can do this using supporting information from other candidates in Table~\ref{tab:qualitative}.

\vspace{-0.25em}

\section{Conclusions}
\label{sec:conclusions}

\vspace{-0.5em}

In this paper we have presented a multi-sentence cross-encoder for performing inference jointly on multiple sentences for tasks like answer sentence selection and fact verification. We have proposed a novel pre-training task to capture paragraph-level semantics. Our experiments on three answer selection and one fact verification datasets show that our pre-trained joint models can outperform pairwise cross-encoders and pre-trained LMs when directly used as joint models.

\bibliography{main}
\bibliographystyle{acl_natbib}

\clearpage
\appendix

\section*{Appendix}

\section{Datasets}
\label{app:datasets}

We present the complete details for all the datasets used in this paper along with links to download them for reproducibility of results.

\subsection{Pre-training Datasets}
We use the Wikipedia\footnote{\url{https://dumps.wikimedia.org/enwiki/20211101/}}, BookCorpus\footnote{\url{https://huggingface.co/datasets/bookcorpusopen}}, OpenWebText \citep{Gokaslan2019OpenWeb} and CC-News\footnote{\url{https://commoncrawl.org/2016/10/news-dataset-available/}} datasets for performing pre-training of our joint transformer models. We do not use the STORIES dataset as it is no longer available for research use \footnote{\url{https://github.com/tensorflow/models/tree/archive/research/lm\_commonsense\#1-download-data-files}}. After decompression and cleaning we obtained 6GB, 11GB, 38GB and 394GB of raw text respectively from the BookCorpus, Wikipedia, OpenWebText and CC-News.

\subsection{Finetuning Datasets}
We evaluate our joint transformers on three AS2 and one Fact Verification datasets. The latter differs from the former in not selecting the best candidate, but rather explicitly using all candidates to predict the target label.
Here are the details of the finetuning datasets that we use for our experiments along with data statistics for each dataset:

\begin{table}[h]
    \centering
    \resizebox{\linewidth}{!}{%
    \begin{tabular}{clccc}
    \toprule
        \textbf{Dataset} &  \textbf{Split} & \textbf{\# Questions} & \textbf{\# Candidates} & \textbf{Avg. \# C/Q} \\
        \midrule
        \multirow{3}{*}{\rotatebox[origin=c]{90}{\small ASNQ}}
        & Train   & 57,242 & 20,377,568  & 356.0 \\
        & Dev     & 1,336  & 463,914    & 347.2 \\
        & Test    & 1,336  & 466,148    & 348.9 \\
        \midrule
        \multirow{3}{*}{\rotatebox[origin=c]{90}{\small WikiQA}}
        & Train   & 2,118  & 20,360     & 9.6 \\
        & Dev     & 122   & 1,126      & 9.2 \\
        & Test    & 237   & 2,341      & 9.9 \\
        \midrule
        \multirow{3}{*}{\rotatebox[origin=c]{90}{\small TREC-QA}}
        & Train   & 1,226  & 53,417     & 43.6 \\
        & Dev     & 69    & 1,343      & 19.5 \\
        & Test    & 68    & 1,442      & 21.2 \\
        \bottomrule
    \end{tabular}
    }
    \caption{Statistics for ASNQ, WikiQA and TREC-QA datasets.}
    \label{tab:as2}
\end{table}

\begin{itemize}[wide, labelwidth=!, labelindent=0pt]
    \item \textbf{ASNQ:} A large-scale AS2 dataset~\cite{garg2019tanda}\footnote{\url{https://github.com/alexa/wqa_tanda}} where the candidate answers are from Wikipedia pages and the questions are from search queries of the Google search engine. ASNQ is a modified version of the Natural Questions (NQ)~\cite{kwiatkowski-etal-2019-natural} dataset by converting it from a machine reading to an AS2 dataset. This is done by labelling sentences from the long answers which contain the short answer string as positive correct answer candidates and all other answer candidates as negatives. We use the dev. and test splits released by ~\citeauthor{soldaini-moschitti-2020-cascade}\footnote{\url{https://github.com/alexa/wqa-cascade-transformers}}.

    \item \textbf{WikiQA:} An  AS2 dataset released by \citeauthor{yang2015wikiqa}\footnote{\url{http://aka.ms/WikiQA}} where the questions are derived from query logs of the Bing search engine, and the answer candidate are extracted from Wikipedia. This dataset has a subset of questions having no correct answers (all-) or having only correct answers (all+). We remove both the all- and all+ questions for our experiments (``clean" setting).
    
    \item \textbf{TREC-QA:} A popular AS2 dataset released by \citeauthor{wang-etal-2007-jeopardy}. For our experiments, we trained on the \textit{train-all} split, which contains more noise but also more question-answer pairs. Regarding the dev. and test sets we removed the questions without answers, or those having only correct or only incorrect answer sentence candidates. This setting refers to the ``clean" setting~\cite{shen-etal-2017-inter}, which is a TREC-QA standard.
    
    \item \textbf{FEVER:} A popular benchmark for fact extraction and verification released by \citeauthor{thorne-etal-2018-fever} The aim is to retrieve evidences given a claim, and then identify whether the retrieved evidences support or refute the claim or if there is not enough information to make a choice. For supporting/refuting a claim, at least one of the retrieved evidences must support/retrieve the claim. Note that the performance on FEVER depends crucially on the retrieval system and the candidates retrieved. For our experiments, we are interested only in the fact verification sub-task and thus we exploit the evidences retrieved by \citeauthor{liu2020kernel} using a BERT-based DocIR\footnote{\url{https://github.com/thunlp/KernelGAT/tree/master/data}}.
    
    \begin{table}[h]
    \centering
    \small
    \begin{tabular}{lccc}
    \toprule
        \textbf{Split} & \textbf{\# Claims} & \textbf{\# Evidences} & \textbf{Avg. \# E/C} \\
        \midrule
        Train   & 145,406    & 722,473 & 4.97 \\
        Dev     & 19,998     & 98,915  & 4.95 \\
        Test    & 19,998     & 98,839  & 4.94 \\ \bottomrule
    \end{tabular}
    \caption{Statistics for the FEVER dataset where evidences has been retrieved using \citep{liu2020kernel}.}
    \label{tab:fever}
    \end{table}    
\end{itemize}

\section{Experimental Setup}
\label{app:experiments}

\subsection{Complete Experimental Details}
Following standard practice, the token ids, positional ids and token type ids are embedded using separate embedding layers, and their sum is fed as the input to the transformer layers.
We use $k{=}5$ for our experiments (following \citeauthor{zhang-etal-2021-joint, tymoshenko-moschitti-2021-strong}), and perform continuous pre-training starting from the RoBERTa-Base checkpoint using a combination of MLM and our MSPP pre-training objective for 100,000 steps with a batch size of 4096. We use a triangular learning rate with 10,000 warmup steps and a peak value of $5*10^{-5}$. We use Adam optimizer with $\beta_1 = 0.9$, $\beta_2 = 0.999$ and $\epsilon = 10^{-8}$. We apply a weight decay of $0.01$ and gradient clipping when values are higher than $1.0$. We set the dropout ratio to $0.1$ and we use two different prediction heads for pre-training: IE$_k$ and AE$_k$. We follow the strategy of ~\cite{devlin2019bert,lan2020albert}, and equally weight the the two pre-training loss objectives: MLM and MSPP.

For evaluation, we fine-tune all models on the downstream AS2 and FEVER datasets: using the same IE$_k$ and AE$_k$ prediction heads exploited in pre-training for AS2 and using either IE$_1$ or AE$_1$ prediction heads for FEVER. We finetune every model with the same maximum sequence length equal to $64 * (k+1) = 384$ tokens. For ASNQ we train for up to 6 epochs with a batch size of 512 and a learning rate of $10^{-5}$ with the same Adam optimizer described above but warming up for only 5000 steps. We do early stopping on the MAP of the development set. For WikiQA and TREC-QA, we created batches of 32 examples and we used a learning equal to $2*10^{-6}$ and 1000 warm up steps. We train for up to 40 epochs again with early stopping on the MAP of the development set. On FEVER, we use a batch size of 64, a learning rate of $10^{-5}$, 1000 warm up steps and we do early stopping checking the Accuracy over the development set. We implemented our code based on HuggingFace's Transformers library~\cite{wolf-etal-2020-transformers}.

\subsection{Baselines}
For AS2, we consider two baselines: (i) pairwise RoBERTa-Base model when used as a cross-encoder for AS2, and (ii) RoBERTa-Base LM when used as a joint model with IE$_k$ and AE$_k$ prediction heads independently for AS2 tasks.

For FEVER, we use several recent baselines from \citeauthor{tymoshenko-moschitti-2021-strong}: (i) GEAR~\cite{zhou-etal-2019-gear}, (ii) KGAT~\cite{liu2020kernel}, (iii) Transformer-XH~\cite{zhao2020transformer-xh}, (iv) joint RoBERTa-Base with IE$_1$ prediction head~\cite{tymoshenko-moschitti-2021-strong}, (v) pairwise RoBERTa-Base when used as a cross-encoder with max-pooling head~\cite{tymoshenko-moschitti-2021-strong}, (vi) pairwise RoBERTa-Base when used as a cross-encoder with weighted-sum head~\cite{tymoshenko-moschitti-2021-strong}.

We used metrics from Torchmetrics~\cite{torchmetrics} to compute MAP, MRR, Precision@1 and Accuracy.

\subsection{Metrics}

The performance of AS2 systems in practical applications is typically~\cite{garg-moschitti-2021-will} measured using the Accuracy in providing correct answers for the questions (the percentage of correct responses provided by the system), also called the Precision-at-1 (P@1). In addition to P@1, we use Mean Average Precision (MAP) and Mean Reciprocal Recall (MRR) to evaluate the ranking produced of the set of candidates by the model.

For FEVER, we measure the performance using Label Accuracy (LA), a standard metric for this dataset, that measures the accuracy of predicting support/refute/neither for a claim using a set of evidences.

\section{Complete Results and Discussion}
\label{app:complete_results}

\begin{table*}[t]
    \centering
    \resizebox{0.8\linewidth}{!}{
    \begin{tabular}{lccccccccccc}
    \toprule
    \multirow{2}{*}{\textbf{Model}} & \multicolumn{3}{c}{\textbf{ASNQ}} & &\multicolumn{3}{c}{\textbf{WikiQA}} & & \multicolumn{3}{c}{\textbf{TREC-QA}} \\ \cmidrule{2-4} \cmidrule{6-8} \cmidrule{10-12}
                       & \textbf{P@1}    & \textbf{MAP}    & \textbf{MRR}   & &\textbf{ P@1}     & \textbf{MAP}     & \textbf{MRR}   & & \textbf{P@1}   & \textbf{MAP}     & \textbf{MRR}     \\
    \midrule
Pairwise RoBERTa-Base   & 61.8   & 66.9   & 73.1  & & 77.1    & 85.3    & 86.5  & & 87.9    & 89.3    & 93.1    \\
Joint RoBERTa-Base ${\rightarrow}$ FT IE$_k$   &  25.2    &  44.0      &  45.6     & & 24.6    & 49.3    & 49.7   && 57.6    & 73.7    & 74.6    \\
Joint RoBERTa-Base ${\rightarrow}$ FT AE$_k$ &   25.4     &    44.8    &    46.2    && 26.4    & 50.6    & 51.1   && 60.9   & 74.6    & 76.7    \\
(\textbf{Ours}) Joint MSPP IE$_k$ ${\rightarrow}$ FT IE$_k$       & 63.9   & 71.3   & 73.1   && \textbf{82.7}    & \textbf{88.5}    & \textbf{89.0}   && \textbf{92.2}    & \textbf{93.5}    & \textbf{95.4} \\
(\textbf{Ours}) Joint MSPP AE$_k$ ${\rightarrow}$ FT AE$_k$   & \textbf{64.3}   & \textbf{71.5}   & \textbf{73.4}   && 82.1    & 87.9    & 88.7   && 91.2    & \textbf{93.5}    & 94.9   \\
\bottomrule
\end{tabular} }
    \caption{Complete results of our joint models for AS2 datasets when \textbf{re-ranking the answer candidates ranked in top-k by Pairwise RoBERTa-Base}. MSPP, FT refer to our pre-training task and finetuning respectively. We indicate the prediction head (IE$_k$/AE$_k$) used for both pre-training and finetuning.}
    \label{tab:app_results_as2_topk}
\end{table*}

\begin{table}[t]
    \centering
    \resizebox{\linewidth}{!}{
    \begin{tabular}{lcccc}
    \toprule
        \textbf{Model} & \textbf{Dev} & \textbf{Test} \\
        \midrule
        GEAR & 70.69 & 71.60 \\
        KGAT with RoBERTa-Base & 78.29 & 74.07 \\
        Transformer-XH & 78.05 & 72.39 \\
        Pairwise BERT-Base & 73.30 & 69.75 \\
        Pairwise RoBERTa-Base + MaxPool & 79.82 & - \\
        Pairwise RoBERTa-Base + WgtSum & 80.01 & - \\
        Joint BERT-Base & 73.67 & 71.01 \\
        Joint RoBERTa-Base + FT IE$_1$ & 79.25 & 73.56 \\
        (\textbf{Ours}) Joint Pre IE$_k$ + FT IE$_1$  & \bfunder{81.21} (0.24)    & \textbf{74.39} \\
        (\textbf{Ours}) Joint Pre IE$_k$ + FT AE$_1$  & \underline{81.10} (0.15)  & 74.25 \\
        (\textbf{Ours}) Joint Pre AE$_k$ + FT IE$_1$  & \underline{81.18} (0.14)  & 73.77  \\
        (\textbf{Ours}) Joint Pre AE$_k$ + FT AE$_1$  & \bfunder{81.21} (0.16)    & 74.13 \\ \midrule
        \multicolumn{3}{c}{\textbf{Methods with larger models and/or sophisticated retrieval}}\\
        DOMLIN++ & 77.48 & 76.60 \\
        DREAM & 79.16 & 76.85 \\
        \bottomrule
    \end{tabular}}
    \caption{Complete Results on FEVER dev and test sets. For our method, prediction heads (IE$_1$/AE$_1$) are only used for finetuning (FT), while for pre-training (Pre) we use the (IE$_k$/AE$_k$) heads. '-' denotes models that are not publicly released and have no reported results on the test split in their published paper. Statistically significant results (T-Test $95\%$) are underlined.}
    \label{tab:app_results_fever}
    \vspace{.5em}
\end{table}

\subsection{Results on AS2 with cascaded pairwise and Joint re-ranker}

Below we present results of evaluating our joint models to re-rank the top-$k$ candidates ranked by the pairwise RoBERTa-Base cross-encoder. Our joint models can significantly improve the P@1, MAP and MRR over the baseline for all datasets. The performance gap stems from questions for which the pairwise RoBERTa model was unable to rank the correct answer at the top position, but support from other candidates in the top-k helped the joint model rank it in the top position.

\subsection{Results on FEVER}
Here we present complete results on the FEVER dataset in Table~\ref{tab:app_results_fever}, by also presenting some additional baselines such as: (i) pairwise BERT-Base cross-encoder~\cite{tymoshenko-moschitti-2021-strong}, (ii) joint BERT-Base cross-encoder with IE$_1$ prediction head, (iii) DOMLIN++~\cite{Stammbach2020eFEVEREA} which uses additional DocIR components and data (MNLI~\cite{williams-etal-2018-broad}) for fine-tuning, (iv) DREAM~\cite{zhong-etal-2020-reasoning} that uses the XL-Net model. Note that comparing our joint models with (iii) and (iv) is unfair since they use additional retrieval components, datasets and larger models. We just include these results here for the sake for completeness. Interestingly, our joint models outperform DREAM and DOMLIN++ on the dev set without using additional retrieval and larger models. 

\subsection{Compute Overhead of Joint Models}
\label{app:number_params}

\mypara{Change in Number of Model Parameters:}
The transformer block of our joint inference model is identical to pre-trained models such as RoBERTa, and contains the exact same number of parameters. Classification heads $IE_1, IE_k$ and $AE_1$ all operate on the embedding of a single token, and are identical to the classification head of RoBERTa ($AE_k$ operates on the concatenation of two token embeddings, and contains double the number of parameters as the RoBERTa). The maximum sequence length allowed for both the models is the same (512). The exact number of parameters of our joint model with $AE_k$ and the RoBERTa model are $124,062,720$ and $124,055,040$ respectively.

\mypara{Change in Inference Latency:} While our joint model provides a longer input sequence to the transformer, it also reduces the number of forward passes that need to be done by a pairwise cross-encoder. A \emph{simplified} latency analysis for AS2 (assuming each sentence has a length $L$): pairwise cross-encoder will need to make $k$ forward passes of the transformer with a sequence of length $2L$ ($q$ with each candidate $c_i$), while our joint model will only need to make $1$ forward pass of the transformer with input length $(k{+}1){\times}L$ ($q$ with $k$ candidates). Transformer self-attention is quadratic in input sequence length, so this should lead to the inference time of out joint model being $\frac{(k+1)^2}{4k}$ times the inference time of the cross-encoder. However, the input embedding layer and the feedforward layers are linear in input sequence length, so this should lead to a reduction in the inference time of our joint model by $\frac{(k+1)}{2k}$ times the inference time of the cross-encoder. Empirically, when we fine-tune for WikiQA on one A100-GPU, we only observe latency increasing from $71s {\rightarrow} 81s$ (increase of only $14.1\%$).

\end{document}